%% file: iclr2025_conference.tex
\documentclass[11pt, a4paper, logo, copyright]
{googledeepmind}

\input{math_commands.tex}

\usepackage[authoryear, sort&compress, round]{natbib}
\usepackage[capitalise]{cleveref}
\usepackage{enumitem}

\usepackage{amsmath}
\usepackage{amssymb}
\usepackage{amsfonts}
\usepackage{titlesec}
\usepackage{graphicx}
\usepackage{paralist}
\usepackage{listings}
\usepackage{enumitem}
\usepackage{authblk}
\usepackage{float}
\usepackage{subcaption}

\usepackage{amsmath}
\usepackage{amssymb}
\usepackage{amsthm}
\usepackage{mathtools}
\theoremstyle{plain}

\theoremstyle{definition}

\newcommand{\pibf}{{\boldsymbol\pi}}

\newcommand{\jr}{\text{JumpReLU}}
\newcommand{\pseudopartial}{\textup{\rmfamily\dh}}

\usepackage{xcolor}
\usepackage[T1]{fontenc}
\usepackage[utf8]{inputenc}
\usepackage[T1]{fontenc}

\title{Resurrecting the Salmon: Rethinking Mechanistic Interpretability with Domain-Specific Sparse Autoencoders}

\author[1]{Charles O'Neill}
\author[1]{Mudith Jayasekara}
\author[1]{Max Kirkby}

\affil[1]{Parsed, London, UK}

\begin{abstract}
Sparse autoencoders (SAEs) decompose large language model (LLM) activations into latent features that reveal mechanistic structure. Conventional SAEs train on broad data distributions, forcing a fixed latent budget to capture only high-frequency, generic patterns. This often results in significant linear ``dark matter'' in reconstruction error and produces latents that fragment or absorb each other, complicating interpretation. We show that restricting SAE training to a well-defined domain (medical text) reallocates capacity to domain-specific features, improving both reconstruction fidelity and interpretability. Training JumpReLU SAEs on layer-20 activations of Gemma-2 models using 195k clinical QA examples, we find that domain-confined SAEs explain up to 20\% more variance, achieve higher loss recovery, and reduce linear residual error compared to broad-domain SAEs. Automated and human evaluations confirm that learned features align with clinically meaningful concepts (e.g., ``taste sensations'' or ``infectious mononucleosis''), rather than frequent but uninformative tokens. These domain-specific SAEs capture relevant linear structure, leaving a smaller, more purely nonlinear residual. We conclude that domain-confinement mitigates key limitations of broad-domain SAEs, enabling more complete and interpretable latent decompositions, and suggesting the field may need to question ``foundation-model'' scaling for general-purpose SAEs.
\end{abstract}

\begin{document}

\maketitle

\section{Introduction}
\label{sec:introduction}

Sparse autoencoders (SAEs) are employed in mechanistic interpretability to decompose the hidden activations of large language models (LLMs) into sparse latent representations that ideally correspond to semantically distinct, causal features. In these models, an SAE learns to reconstruct input activations using a fixed latent budget and a sparsity constraint, thereby forcing the network to represent each activation as a sparse linear combination of latent directions. Prior work has applied SAEs to models such as GPT‑4 and Claude to identify interpretable circuits and uncover the features underlying model behaviour \citep{cunningham2023sparse, bricken2023towards}.

Several studies indicate that when SAEs are trained on broad data distributions, a number of limitations arise. When trained on a broad dataset, the fixed latent budget forces the SAE to capture only the most common, high-frequency patterns, leaving little capacity for fine-grained, domain‐specific features. For instance, despite increasing coverage of concepts as the capacity of the SAE is increased, there is evidence that even in the largest SAEs, the set of features uncovered is an incomplete description of the model’s internal representations \citep{transformercircuitsScalingMonosemanticity}.\footnote{For example, \citet{transformercircuitsScalingMonosemanticity} confirmed that Claude 3 Sonnet can list all of the London boroughs when asked, and in fact can name tens of individual streets in many of the areas. However, features corresponding to approximately 60\% of the boroughs were found in the 34M SAE. This suggests that the model contains many more features than they have found, which may be extracted with even larger SAEs.} A large portion of SAE error is linearly predictable from input activations, suggesting a multitude of unlearned features \citep{engels2024decomposing}. This residual error leads to substantial downstream substitution loss when the SAE reconstruction is reinserted into the language model \citep{gao2024scaling}.

\begin{figure}
    \centering
    \includegraphics[width=0.8\linewidth]{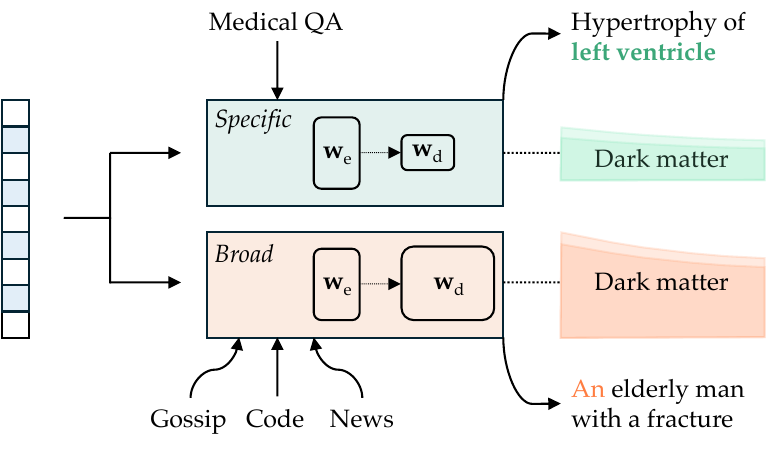}
    \caption{Pareto curves of fraction of variance explained across different models (\texttt{gemma-2-2b} and \texttt{gemma-2-9b} from \citet{gemmateam2024gemma2improvingopen}), SAE widths, and sparsities, comparing our SAEs to the GemmaScope \citep{lieberum2024gemma} SAEs.}
    \label{fig:frac_variance_explained}
\end{figure}

\color{black}

Studies also demonstrate that SAEs can exhibit feature splitting, where a latent that should represent a single interpretable concept instead fragments into multiple, more specific latents. Moreover, feature absorption occurs when token‐aligned latents ``absorb'' an expected feature direction, causing the intended latent to fail to activate in some contexts and reducing overall recall \citep{chanin2024absorption}. \citet{till_2024_saes_true_features} argues that the L1 regularisation used to enforce sparsity can drive the SAE to learn common combinations of features rather than the atomic ``true features'' that causally mediate the model’s computations, although this has been ameliorated somewhat with newer architectures and optimisation techniques \citep{gao2024scaling, rajamanoharan2024improving, rajamanoharan2024jumping}. Recent work has also shown that SAEs trained on the same model and data, differing only in their random initialisation, learn substantially different feature sets \citep{paulo2025sparse}, indicating that SAE decomposition is not unique but rather a pragmatic artifact of training conditions.

Collectively, these findings indicate that the conventional "more is better" scaling paradigm for language models does not effectively translate to mechanistic interpretability using SAEs. Broad-domain training produces latent representations that are generic, inconsistent, and vulnerable to issues such as nonlinear error and feature absorption. We hypothesise that narrowing the input domain could encourage the SAE's fixed latent capacity to selectively represent only high-fidelity, task-relevant features. This reallocation of capacity is expected to reduce downstream substitution error and yield latent representations that more accurately reflect the causal circuitry of the target model \citep{gao2024scaling, chanin2024absorption}. 

In this paper, we argue that mechanistic interpretability via SAEs requires a shift from broad-domain scaling to domain-specific training. To support this claim, we conduct a thorough study of SAEs applied to medical text data, and demonstrate empirically and theoretically the benefits of training domain-specific SAEs. Specifically, we train JumpReLU SAEs with the same capacity as various GemmaScope SAEs \citep{lieberum2024gemma} and determine how training on a well-defined but narrow domain addresses many, if not all, of the concerns with SAEs outlined above through \textit{unsupervised evaluation}. We then demonstrate that the features learned by our domain-specific SAEs are not only much more specific, but more \textit{interpretable} than GemmaScope features pertaining to medicine. Finally, we show how the minimal linear and nonlinear variance in our domain-specific SAEs leads us to conclude that reasonably sized domain-specific SAEs can truly learn all required features for said domain.


Taken together, this paper provides evidence that SAEs applied to specific domains are the most promising direction forward for the current paradigms of mechanistic interpretability.

\section{Background}
\label{sec:background}

\subsection{Sparse Autoencoder (SAE) Architectures}
\label{sec:sparse-autoencoder}

Our goal is to decompose a model's activation $\mathbf{x} \in \mathbb{R}^n$ into a sparse linear combination of learned feature directions. Intuitively, we express:
\begin{equation}
\mathbf{x} \approx \mathbf{x}_0 + \sum_{i=1}^M f_i(\mathbf{x})\, \mathbf{d}_i,
\end{equation}
where $\mathbf{d}_i$ are $M \gg n$ latent unit-norm feature directions and the coefficients $f_i(\mathbf{x})\geq 0$ denote the activation strength for each feature.

In a sparse autoencoder (SAE), the encoder and decoder are defined as:
\begin{align}
\mathbf{f}(\mathbf{x}) &:= \sigma\Big(\mathbf{W}_{\mathrm{enc}}\,\mathbf{x} + \mathbf{b}_{\mathrm{enc}}\Big), \label{eq:encoder}\\[1ex]
\hat{\mathbf{x}}(\mathbf{f}) &:= \mathbf{W}_{\mathrm{dec}}\,\mathbf{f} + \mathbf{b}_{\mathrm{dec}}, \label{eq:decoder}
\end{align}
where $\mathbf{f}(\mathbf{x})\in \mathbb{R}^M$ is a sparse, non-negative vector, and the columns of $\mathbf{W}_{\mathrm{dec}}$, denoted $\mathbf{d}_i$, form the dictionary. For convenience, we define the encoder's pre-activations as:
\begin{equation}
\pibf(\mathbf{x}) := \mathbf{W}_{\mathrm{enc}}\,\mathbf{x} + \mathbf{b}_{\mathrm{enc}}.
\label{eq:preactivation}
\end{equation}

While standard activations such as ReLU \citep{bricken2023towards, transformercircuitsScalingMonosemanticity} and TopK \citep{gao2024scaling} are commonly used, in our work we adopt the \emph{JumpReLU} activation function \citep{rajamanoharan2024jumping, erichson2019jumprelu}:
\begin{equation}
\jr_\theta(z) := z\,H(z-\theta),
\label{eq:jump-relu}
\end{equation}
where $H(z)$ is the Heaviside step function and $\theta\in\mathbb{R}_+$ is a learnable threshold. The additional parameter $\theta$ enables the network to decide whether a feature is active prior to estimating its magnitude.

\subsection{Loss Functions for SAEs}
\label{sec:loss-functions}

A typical loss function for language model SAEs is formulated as:
\begin{equation}
\mathcal{L}(\mathbf{x}) := \underbrace{\|\mathbf{x} - \hat{\mathbf{x}}(\mathbf{f}(\mathbf{x}))\|_2^2}_{\mathcal{L}_{\mathrm{reconstruct}}} + \underbrace{\lambda\,S\big(\mathbf{f}(\mathbf{x})\big)}_{\mathcal{L}_{\mathrm{sparsity}}} + \mathcal{L}_{\mathrm{aux}},
\label{eq:typical-loss}
\end{equation}
where $S$ is a sparsity penalty such as L1 and $\lambda$ balances reconstruction fidelity and sparsity. 

A JumpReLU SAE modifies the standard sparse autoencoder architecture (Equations~\ref{eq:encoder} and \ref{eq:decoder}) by replacing the usual activation function with a JumpReLU. In this model, the encoder is defined as:
\begin{equation}
\mathbf{f}(\mathbf{x}) := \mathrm{JumpReLU}_{\boldsymbol{\theta}}\Big(\mathbf{W}_{\mathrm{enc}}\,\mathbf{x} + \mathbf{b}_{\mathrm{enc}}\Big),
\end{equation}
where \(\boldsymbol{\theta} \in \mathbb{R}_+^M\) is a vector of positive thresholds, one per feature. Unlike the standard ReLU, this extra parameter \(\boldsymbol{\theta}\) sets a minimum required value for each encoder preactivation to be considered active. Thus, the JumpReLU SAE clearly separates the decision of whether a feature is active from the estimation of its magnitude.

In JumpReLU SAEs, we employ an L0 sparsity penalty:
\begin{equation}
\mathcal{L}(\mathbf{x}) = \|\mathbf{x} - \hat{\mathbf{x}}(\mathbf{f}(\mathbf{x}))\|_2^2 + \lambda\,\|\mathbf{f}(\mathbf{x})\|_0.
\label{eq:jr-loss-full}
\end{equation}
Since the L0 norm simply counts non-zero entries, it can be expressed via the Heaviside step function:
\begin{equation}
\|\mathbf{f}(\mathbf{x})\|_0 = \sum_{i=1}^{M} H\Big(\pi_i(\mathbf{x}) - \theta_i\Big),
\label{eq:l0-step-function-repeat}
\end{equation}
where $\pi_i(\mathbf{x})$ is the $i^{\mathrm{th}}$ component of $\pibf(\mathbf{x})$.

\subsection{Straight-Through Estimators (STEs)}
\label{sec:ste-training}

The discontinuities inherent in the L0 penalty and the threshold parameter $\theta$ result in zero gradients under standard backpropagation. To overcome this, we adopt \emph{straight-through estimators} (STEs) that define pseudo-derivatives for the non-differentiable functions \citep{bengio2013estimating}.

Specifically, we define:
\begin{align}
\frac{\pseudopartial}{\pseudopartial \theta}\jr_\theta(z) &:= -\frac{\theta}{\varepsilon}\, K\!\Big(\frac{z-\theta}{\varepsilon}\Big), \label{eq:jr-ste}\\[1ex]
\frac{\pseudopartial}{\pseudopartial \theta} H(z-\theta) &:= -\frac{1}{\varepsilon}\, K\!\Big(\frac{z-\theta}{\varepsilon}\Big), \label{eq:step-ste}
\end{align}
where $K$ is a kernel function (e.g. the rectangle function defined as $\mathrm{rect}(z):=H(z+1/2)-H(z-1/2)$) and $\varepsilon>0$ is a small bandwidth parameter.

These STEs allow gradient information to pass through the discontinuities. Importantly, they can be interpreted as yielding a kernel density estimation (KDE) for the true gradient (see Appendix~\ref{app:ste-kde-connection}). 


\section{Methodology}

In this section, we outline our experimental approach. We begin by describing the construction of our medical text dataset, then detail the training procedure for our SAEs, and finally introduce the evaluation metrics used to assess the quality of these SAEs compared to GemmaScope.

\subsection{Dataset}

Our first step is to create a domain-specific corpus that captures the essential features relevant to clinical tasks such as differential diagnosis. To achieve this, we combine multiple publicly available datasets, summarised in Table~\ref{tab:medical_datasets}.

\begin{table}[h]
\scriptsize
\centering
\begin{tabular}{lccp{7cm}}
\toprule
\textbf{Dataset} & \textbf{Examples} & \textbf{Type} & \textbf{Description \& Citation} \\
\midrule
MedQA & 10,200 & Multiple Choice & USMLE questions from professional board exams \citep{jin2020diseasedoespatienthave}. \\
MedMCQA & 183,000 & Multiple Choice & Questions from real-world medical entrance exams \citep{pal2022medmcqa}. \\
MMLU College Medicine & 173 & Multiple Choice & College-level medical knowledge questions \citep{hendrycks2021measuringmassivemultitasklanguage}. \\
MMLU Clinical Knowledge & 265 & Multiple Choice & Questions evaluating clinical concepts and practices. \\
MMLU Professional Medicine & 272 & Multiple Choice & Questions assessing professional medical knowledge. \\
PubMedQA & 450 & Question Answering & Biomedical Q\&A with research questions, abstracts (without conclusions), long answers, and a yes/no/maybe summary \citep{jin2019pubmedqadatasetbiomedicalresearch}. \\
\bottomrule
\end{tabular}
\caption{Summary of the medical datasets used in our study.}
\label{tab:medical_datasets}
\end{table}

Each source dataset was standardised into a uniform format, preserving the question text, multiple-choice options, correct answers, and, when available, explanations and contextual details. For PubMedQA entries, we also include the relevant medical context and detailed answer explanations. All fields are concatenated into a single string per example. Preserving a uniform structure allows us to remove the need to learn a significant number of features that correspond simply to the structure of the text rather than its contents.

The final combined dataset comprises approximately 195,000 examples and roughly 50 million tokens (as determined by the Gemma-2 tokeniser \citep{gemmateam2024gemma2improvingopen}). The complete dataset is publicly available on HuggingFace as \href{https://huggingface.co/datasets/charlieoneill/medical-qa-combined}{irisai/medical-qa-combined}.

With our unified dataset in hand, we now describe the training setup used to extract latent features from large language models.

\subsection{Training}

We trained Sparse Autoencoders (SAEs) on the post-MLP output residual stream (layer 20) of both Gemma-2-2b and Gemma-2-9b models. For each model-width combination, we employed a JumpReLU architecture with a bandwidth parameter of $0.001$, trained using MSE reconstruction loss plus a quadratic penalty on deviations from the target sparsity level. Training was conducted using the Adam optimiser with a learning rate of $7\text{e}-5$ and $\beta$ parameters $(0.0, 0.999)$, following the implementation in the original JumpReLU paper \citep{rajamanoharan2024jumping}. We normalised all activation vectors to have unit mean squared norm before training, following the recommendations in \citet{cunningham2023sparse}, and scaled the dictionary weights appropriately after training.

Our activation buffer implementation maintained approximately $30\,000$ contexts (each of length 1,024) of activations in memory at any given time, refreshing the buffer when it became half empty. The buffer collected activations by processing text in batches of 4 contexts and yielded training batches of 2,048 tokens. We used the \texttt{medical-qa-combined} dataset for training.

For Gemma-2-2b, we trained SAEs with dictionary sizes of $2^{14}$ and $2^{16}$ features. Target L0 sparsities (average number of active features per activation vector) were set to 20 for both dictionary sizes. For Gemma-2-9b, we explored a broader range of dictionary sizes: $2^{14}$, $2^{15}$, $2^{16}$, $2^{17}$, $2^{18}$, $2^{19}$, and $2^{20}$ features. Target L0 sparsities were scaled with dictionary width, ranging from 20 to approximately 650.

All models were trained for 49 million tokens with a linear learning rate warmup over the first 1,000 steps and sparsity warmup over the first 5,000 steps. We employed gradient clipping with a maximum norm of 1.0 to ensure training stability. The sparsity penalty coefficient was set to 1.0 throughout training, with the quadratic penalty term scaled by the ratio of actual to target sparsity. Training was conducted using mixed precision (\texttt{bfloat16}) for computational efficiency, with model weights stored in \texttt{float32} precision.

\subsection{Evaluation}

After training our SAEs, we evaluate their performance through a series of quantitative and qualitative metrics. 

\paragraph{Unsupervised Metrics}
We evaluate the quality of trained SAEs using three unsupervised metrics: the L0 sparsity, the fraction of variance explained, and the loss recovered. The L0 sparsity is defined as the expected number of active features per input:
\begin{equation}
    \mathbb{E}_{\mathbf{x} \sim \mathcal{D}}\big[\|\mathbf{f}(\mathbf{x})\|_0\big].
\end{equation}
Reconstruction fidelity is measured by the fraction of variance explained:
\begin{equation}
1 - \frac{\mathrm{Var}\big(\mathbf{x} - \hat{\mathbf{x}}(\mathbf{f}(\mathbf{x}))\big)}{\mathrm{Var}(\mathbf{x})},
\end{equation}
where \(\mathrm{Var}(\mathbf{x})\) denotes the sum of variances across all dimensions. In addition, we assess reconstruction quality via the loss recovered metric. Let \(\operatorname{CE}(\phi)\) be the average cross-entropy loss of the language model when a function \(\phi: \mathbb{R}^n \rightarrow \mathbb{R}^n\) is spliced into the model at the SAE insertion point. If \(\hat{\mathbf{x}} \circ \mathbf{f}\) is the autoencoder function, \(\zeta: \mathbf{x} \mapsto \mathbf{0}\) the zero-ablation function, and \(\mathrm{Id}: \mathbf{x} \mapsto \mathbf{x}\) the identity function, then the loss recovered is given by:
\[
1 - \frac{\operatorname{CE}(\hat{\mathbf{x}} \circ \mathbf{f}) - \operatorname{CE}(\mathrm{Id})}{\operatorname{CE}(\zeta) - \operatorname{CE}(\mathrm{Id})}.
\]
By definition, a SAE that always outputs the zero vector achieves a loss recovered of \(0\%\), while perfect reconstruction yields \(100\%\).

Baseline SAEs trained with an L1 sparsity penalty tend to underestimate feature activations, a phenomenon known as shrinkage. This bias leads to reconstructions whose norms are lower than those of the inputs. To quantify shrinkage, we define the relative reconstruction bias \(\gamma\) as the optimal multiplicative factor that minimises the L2 reconstruction loss:
\[
\gamma := \underset{\gamma'}{\arg \min}\; \mathbb{E}_{\mathbf{x} \sim \mathcal{D}}\Big[\Big\|\frac{\hat{\mathbf{x}}_{\mathrm{SAE}}(\mathbf{x})}{\gamma'} - \mathbf{x}\Big\|_2^2\Big].
\]
An unbiased SAE satisfies \(\gamma=1\), whereas \(\gamma<1\) indicates shrinkage. The derivation of the analytical solution to $\gamma$ in this metric is provided in Appendix \ref{app:reconstruction-bias}.

\paragraph{Interpretability}

Beyond raw reconstruction metrics, it is important to understand whether the latent features are meaningful. We evaluate interpretability by generating explanations for each feature and measuring their fidelity.

We evaluate feature interpretability using the method of \citet{paulo2024automatically}, which involves two stages: (1) generating an interpretation for each feature and (2) assessing the fidelity of that interpretation to the network's true behaviour. If a feature does not support a clear interpretation, the auto-interpretability pipeline will return low evaluation scores. These evaluations can indicate when an SAE produces generally low-quality feature decompositions.

\textit{Generating interpretations.} For each feature, we uniformly sample examples from each activation decile to ensure that explanations are robust across both strong and weak activations. In every example, we mark the tokens with maximum activation using designated delimiters and report their activation magnitudes. In addition, we compute the logit weights via the path expansion \(W_{U}W_{D}[f]\) (where \(W_U\) is the model unembedding matrix and \(W_{D}[f]\) is the decoder direction for feature \(f\)). The top promoted tokens from this expansion capture the feature’s causal effects, thereby sharpening the resulting explanation—an approach equivalent to using a logit lens \citep{bloom2024understandingfeatureslogitlens}.

\textit{Assessing faithfulness.} We measure faithfulness by treating the interpretation as a classifier that predicts whether a feature will activate in a given context. A faithful interpretation should exhibit both high recall (capturing most activating text) and high precision (distinguishing between activating and non-activating text). We employ two methods.

The first is detection: we prompt a language model to determine if an entire sequence activates a given SAE latent based on its interpretation. By including both activating and non-activating contexts, this method evaluates precision and recall without requiring token-level localisation, and it leverages token probabilities to gauge classification confidence. The second is embedding. Here, the interpretation acts as a query to retrieve contexts where the feature is active. We embed both activating and non-activating contexts using an encoding transformer and use the similarity between the query and these contexts to classify them. The resulting classifier is evaluated using the AUROC.

Further details on our automated interpretability pipeline are provided in Appendix~\ref{app:autointerp}.

\paragraph{Dark Matter}

Inspired by the framework of \citet{engels2024decomposing}, we further analyse the reconstruction error of our sparse autoencoders (SAEs) by quantifying linear predictability. For an input activation \( \mathbf{x} \), the SAE produces a reconstruction \(\hat{\mathbf{x}} = \mathrm{SAE}(\mathbf{x})\), yielding an error defined as:
\[
\operatorname{SaeError}(\mathbf{x}) = \mathbf{x} - \mathrm{SAE}(\mathbf{x}).
\]
We hypothesise that a significant portion of this error arises from unlearned, linearly structured features -- ``dark matter''.

To probe this hypothesis, we adopt a three-fold approach. First, we assess the linear predictability of the error norm by learning an optimal scalar probe that maps \( \mathbf{x} \) to the squared error norm \(\|\operatorname{SaeError}(\mathbf{x})\|_2^2\). Formally, we solve for:
\[
\boldsymbol{a}^* = \arg\min_{\boldsymbol{a} \in \mathbb{R}^{d+1}} \left\| \boldsymbol{a}^T \cdot [\mathbf{x};1] - \|\operatorname{SaeError}(\mathbf{x})\|_2^2 \right\|_2^2,
\]
where the augmentation \([\mathbf{x};1]\) accounts for a bias term, and the quality of the fit is measured by the coefficient of determination \(R^2\).

Second, we predict the full error vector by seeking an optimal linear transformation that maps the activation \( \mathbf{x} \) to \(\operatorname{SaeError}(\mathbf{x})\). Specifically, we solve:
\[
\boldsymbol{b}^* = \arg\min_{\boldsymbol{b} \in \mathbb{R}^{(d+1) \times d}} \|\boldsymbol{b} \cdot [\mathbf{x};1] - \operatorname{SaeError}(\mathbf{x})\|_2^2.
\]
The average \(R^2\) computed across activation dimensions quantifies the extent to which the residual error is linearly predictable. A higher \(R^2\) means the residual is more linearly structured -- suggesting the SAE missed certain linear features that remain in the error.

Finally, we evaluate the nonlinear fraction of variance unexplained (FVU) by examining how well the combination of the SAE reconstruction and the linear error prediction accounts for the original activation. Defining:
\[
\tilde{\mathbf{x}} = \mathrm{SAE}(\mathbf{x}) + \boldsymbol{b}^* \cdot \mathbf{x},
\]
we compute:
\[
\mathrm{FVU}_{\text{nonlinear}} = 1 - R^2\left(\mathbf{x}, \tilde{\mathbf{x}}\right).
\]
This metric captures the remaining variance in the activations that is not explained by the sum of the SAE output and the optimal linear prediction of its error. A \emph{larger} FVU\(_{\mathrm{nonlinear}}\) indicates that even after accounting for a linear error term, more variance in \(\mathbf{x}\) remains unexplained, i.e.\ the residual has a greater \emph{nonlinear} component.

For all analyses, activations are extracted from transformer layer 20 after filtering to include only tokens beyond a fixed position within each context, thereby mitigating potential confounds due to token position \citep{lieberum2024gemma}.

\section{Results}
Here we present the results of our analysis comparing domain-specific SAEs with foundational SAEs, across reconstruction fidelity,  interpretability and dark matter analysis.

\begin{figure}[H]
    \centering
    \includegraphics[width=0.98\linewidth]{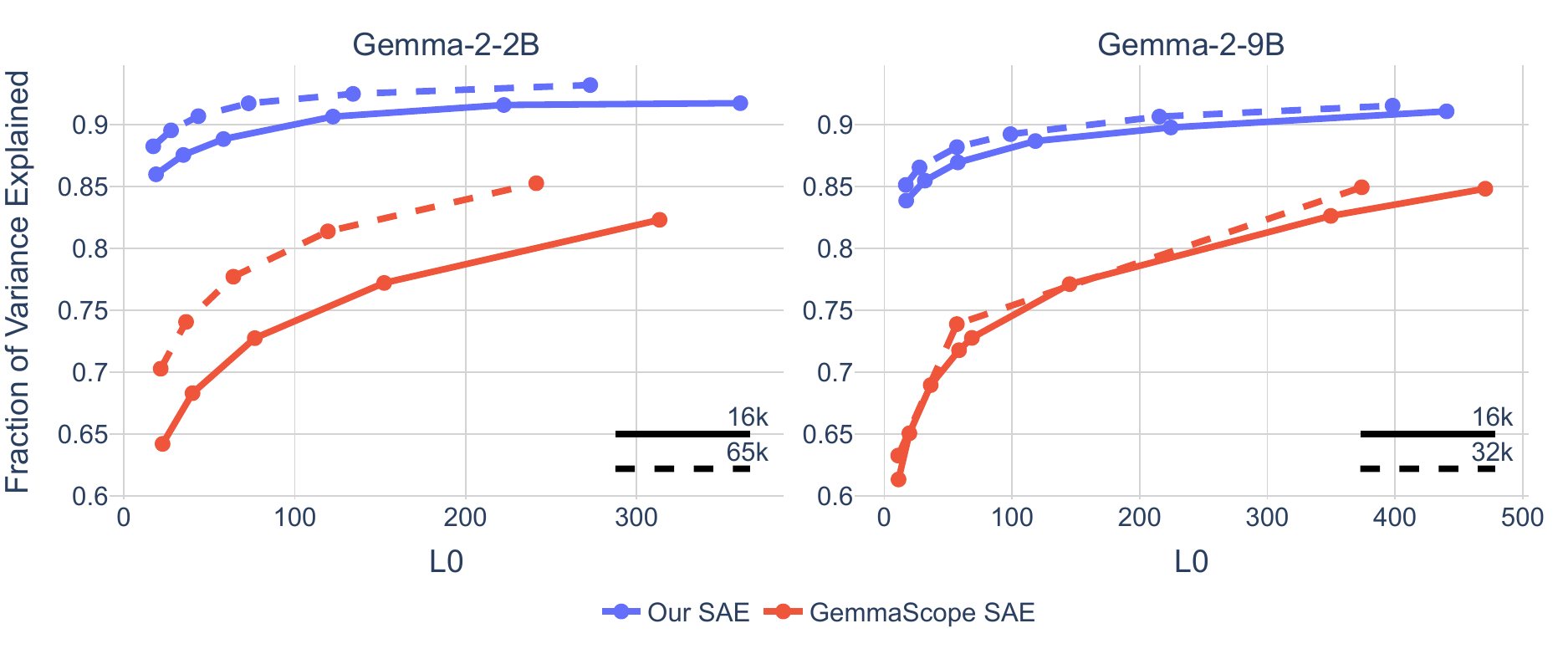}
    \caption{Pareto curves of fraction of variance explained across different models (\texttt{gemma-2-2b} and \texttt{gemma-2-9b} from \citet{gemmateam2024gemma2improvingopen}), SAE widths, and sparsities, comparing our SAEs to the GemmaScope \citep{lieberum2024gemma} SAEs.}
    \label{fig:frac_variance_explained}
\end{figure}

\color{black}

\subsection{Unsupervised Evaluations}

The fraction of variance explained for our SAEs is consistently approximately 15-20\% higher than the accompanying GemmaScope SE, across all L0s (Figure~\ref{fig:frac_variance_explained}). Similarly, the loss recovered (compared to the original Gemma model) when substituting in our SAE activations is also consistently higher (Figure~\ref{fig:loss_recovered}).

We show additional evaluation metrics in Appendix~\ref{app:additional_sae_evaluations}. We find all SAEs, both ours and GemmaScope, to have close to perfect relative reconstruction bias, meaning there is minimal shrinkage (Appendix C, Figure~\ref{fig:relative_reconstruction_bias}). We also show improvements of our SAEs over GemmaScope in cosine similarity between the model activations and SAE reconstructions (Appendix F, Figure \ref{fig:cossim}).

\begin{figure}
    \centering
    \includegraphics[width=0.98\linewidth]{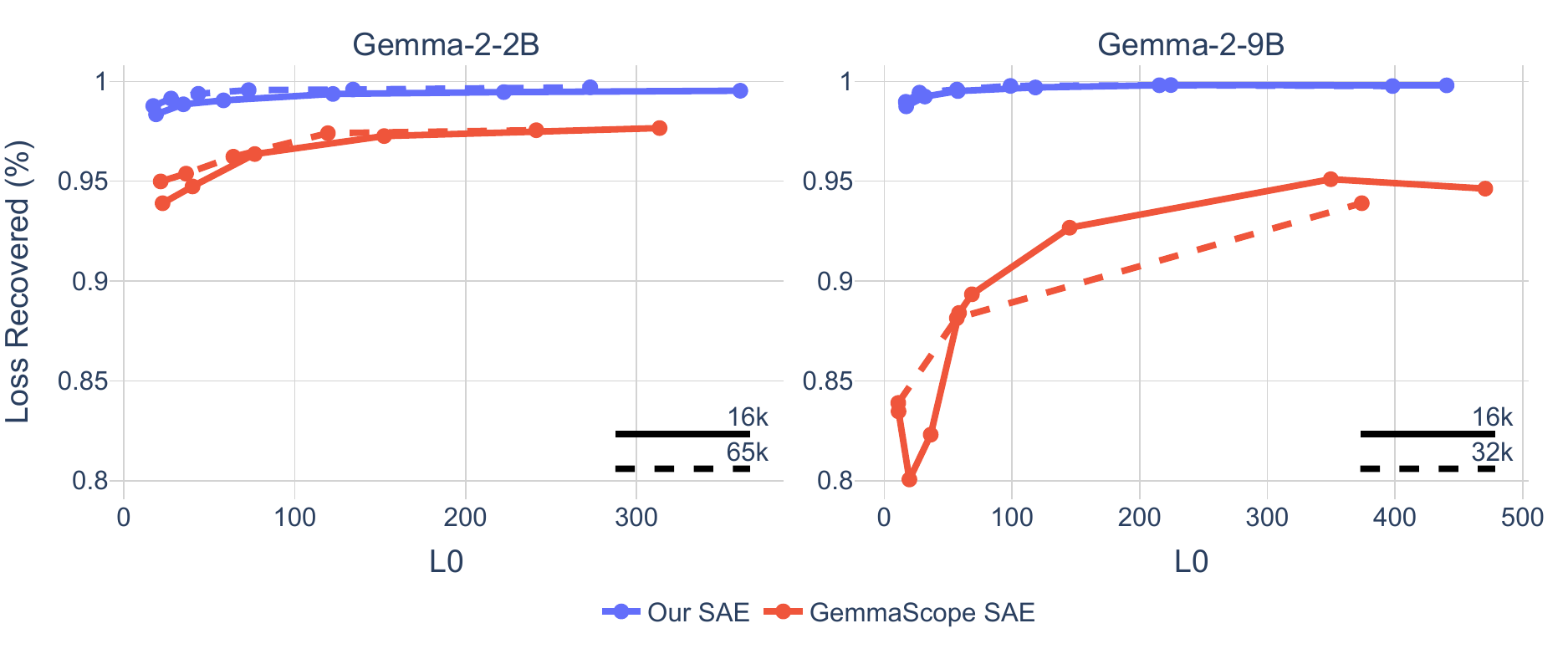}
    \caption{Pareto curves of loss recovered when substituting SAE reconstructions into the model, across different models (\texttt{gemma-2-2b} and \texttt{gemma-2-9b}), SAE widths, and sparsities, comparing our SAEs to the GemmaScope SAEs.}
    \label{fig:loss_recovered}
\end{figure}

\subsection{Interpretability}

\begin{table}[htbp]
\scriptsize
\centering
\begin{tabular}{p{0.2\textwidth} p{0.1\textwidth} p{0.60\textwidth}}
\toprule
\textbf{Feature Explanation} & \textbf{F1 Score} & \textbf{Representative Examples} \\
\midrule
Taste Sensations & 0.94 & 
"smell, or \textcolor{green}{taste} of food, instead of the previous normal salivary secretion by the parotid gland" \\[5pt]
& & "Loss of \textcolor{green}{taste} sensation in anterior 2/3 of tongue due to chorda tympani involvement" \\
\midrule
Specificity in Diagnostic Testing & 0.94 & 
"High \textcolor{green}{specificity} of the test ensured minimal false positives" \\[5pt]
& & "The initial laboratory test was notably \textcolor{green}{specific} in confirming the diagnosis" \\
\midrule
Infectious Mononucleosis & 1.00 & 
"Positive Paul Bunnell test confirmed \textcolor{green}{mononucleosis} in the patient" \\[5pt]
& & "Detection of heterophile antibodies was indicative of \textcolor{green}{mononucleosis}" \\
\midrule
Enzyme Catalase in Differentiating Bacterial Species & 0.91 & 
"Staphylococci were determined to be \textcolor{green}{catalase} positive using the standard test" \\[5pt]
& & "The \textcolor{green}{catalase} test differentiated between bacterial species based on enzyme activity" \\
\midrule
Negation or Lack of Association & 0.97 & 
"Water remains \textcolor{green}{unchanged} in a burn patient, indicating no effect on fluid balance" \\[5pt]
& & "The drug did \textcolor{green}{not alter} the resting membrane potential of the cells" \\
\midrule
Femur in Orthopedic Injuries & 1.00 & 
"An elderly woman sustained a fracture of the \textcolor{green}{femur} following a fall" \\[5pt]
& & "A fracture at the \textcolor{green}{femoral neck} was observed on X-ray imaging" \\
\midrule
Atria and Ventricles in Cardiovascular Anatomy & 0.93 & 
"ECG revealed enlargement of the \textcolor{green}{right atrium} suggesting atrial dilation" \\[5pt]
& & "Echocardiography showed hypertrophy of the \textcolor{green}{left ventricle} consistent with chronic pressure overload" \\
\midrule
Conjunctions/Prepositions for Exceptions/Inclusions & 0.92 & 
"All diagnostic criteria were met, \textcolor{green}{except} for one minor finding" \\[5pt]
& & "The patient improved \textcolor{green}{although} some laboratory values remained borderline" \\
\midrule
Hair Cells in Auditory System & 0.91 & 
"Damage to the \textcolor{green}{hair cells} in the cochlea can lead to significant hearing loss" \\[5pt]
& & "The study focused on the function of the \textcolor{green}{outer hair cells} in auditory transduction" \\
\midrule
Lapse Feature & 0.97 & 
"Imaging revealed a clear \textcolor{green}{prolapse} of the mitral valve, associated with Marfan syndrome" \\[5pt]
& & "A uterine \textcolor{green}{prolapse} was noted during the pelvic examination" \\
\bottomrule
\end{tabular}
\caption{Clinically relevant interpretable features with their F1 scores and representative examples. Activating tokens are highlighted in green.}
\label{tab:clinical_features}
\end{table}

\begin{table}[htbp]
\scriptsize
\centering
\begin{tabular}{p{0.2\textwidth} p{0.1\textwidth} p{0.6\textwidth}}
\toprule
\textbf{Feature Explanation} & \textbf{F1 Score} & \textbf{Representative Examples} \\
\midrule
Image and Imaging in Medical Contexts & 0.913 & 
“Periapical \textcolor{green}{image} reveals bone destruction similar to periodontal disease around the lateral incisor” \\[5pt]
& & “Acute leukemia: \textcolor{green}{image} shows blast cells, suggesting acute leukemia” \\
\midrule
The term ``often'' & 0.980 & 
“Nausea, vomiting, and abdominal guarding are \textcolor{green}{often} seen in the patient” \\[5pt]
& & “Colicky abdominal pain is \textcolor{green}{often} accompanied by the passage of blood and mucus per rectum” \\
\midrule
Frequent Use of “get” and Its Variations & 0.990 & 
“She notes that her symptoms \textcolor{green}{get} much worse when exposed to sunlight” \\[5pt]
& & “The fragments \textcolor{green}{get} nipped between the condyles of tibia and femur, preventing full extension” \\
\midrule
``Overall'' for Summarisation & 0.913 & 
“When one mole of O\textsubscript{2} binds, it causes a shift in the \textcolor{green}{overall} conformation of the protein” \\[5pt]
& & “Adenocarcinoma of the breast now has an \textcolor{green}{overall} 5-year survival rate of 60–70\%” \\
\midrule
Articles ``a'' and ``an'' & 0.943 & 
“\textcolor{green}{A} left shift was noted in the complete blood count” \\[5pt]
& & “\textcolor{green}{An} irregular mass protruding from the vaginal wall was observed on examination” \\
\midrule
Use of “nothing,” “anything,” and “something” & 0.913 & 
``The patient reports that there is \textcolor{green}{nothing} abnormal on examination'' \\[5pt]
& & ``She insists that there is \textcolor{green}{something} wrong in her head despite normal tests'' \\
\midrule
Frequent References to “block,” “blocking,” and “obstruction” & 0.936 & 
“The electrical impulse is \textcolor{green}{blocked}, necessitating an accessory pacemaker” \\[5pt]
& & “A large bolus of air caused an \textcolor{green}{obstruction} in the right atrium and ventricle” \\
\midrule
Frequent Use of “used” and Its Variations & 0.990 & 
“Penicillin 250 mg 12-hourly may be \textcolor{green}{used} if the patient is allergic to penicillin” \\[5pt]
& & “Contrast \textcolor{green}{used} for MRI provided enhanced visualization of the lesion” \\
\midrule
Frequent References to Statistical Measures (Mean, Average, Age) & 0.969 & 
“The \textcolor{green}{average} age of onset was 60 years with a standard deviation of 5” \\[5pt]
& & “The \textcolor{green}{mean} corpuscular volume was calculated to be 90 µm³” \\
\bottomrule
\end{tabular}
\caption{Clinically relevant interpretable features for the GemmaScope 32k width SAE with their F1 scores and representative examples. Activating tokens are highlighted in green.}
\label{tab:gemmascope_features}
\end{table}

\begin{figure}
    \centering
    \includegraphics[width=0.98\linewidth]{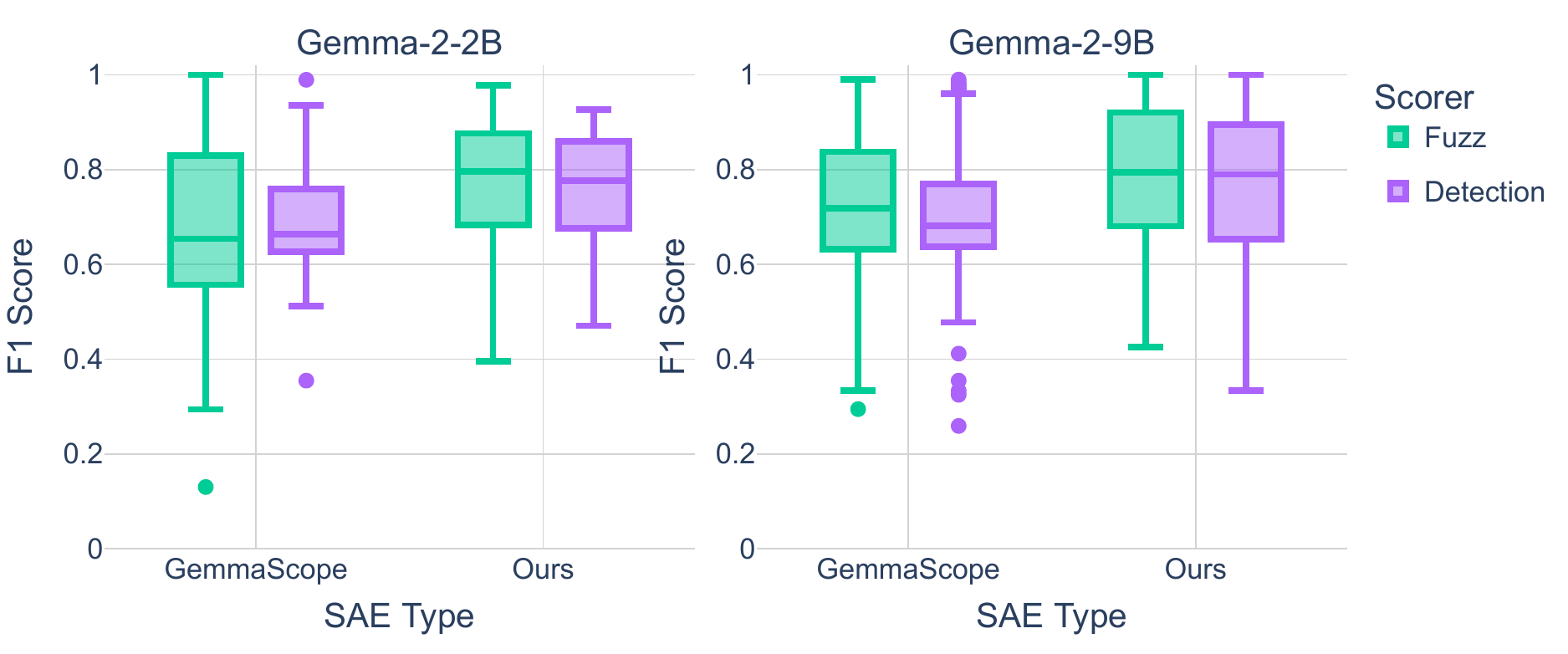}
    \caption{F1 scores of detection and fuzzing evaluations used automated interpretability on our SAE and GemmaScope SAE, on Gemma-2-9B (width 65k latents) and Gemma-2-9B (32k latents).}
    \label{fig:f1_score_comparison-2b-9b}
\end{figure}

Some analysis and visualisation of feature similarities, both within SAEs and between ours and GemmaScope, are given in Appendix~\ref{app:feature-sim}.

\subsection{Dark Matter}

To gain insight into \emph{what} the SAE fails to reconstruct, we follow the ``dark matter'' framework of \citet{engels2024decomposing} and ask: \emph{How much of the SAE's residual error is itself just unlearned \textbf{linear} structure, and how much is fundamentally nonlinear?}

\begin{figure}[t]
    \centering
    \includegraphics[width=0.95\linewidth]{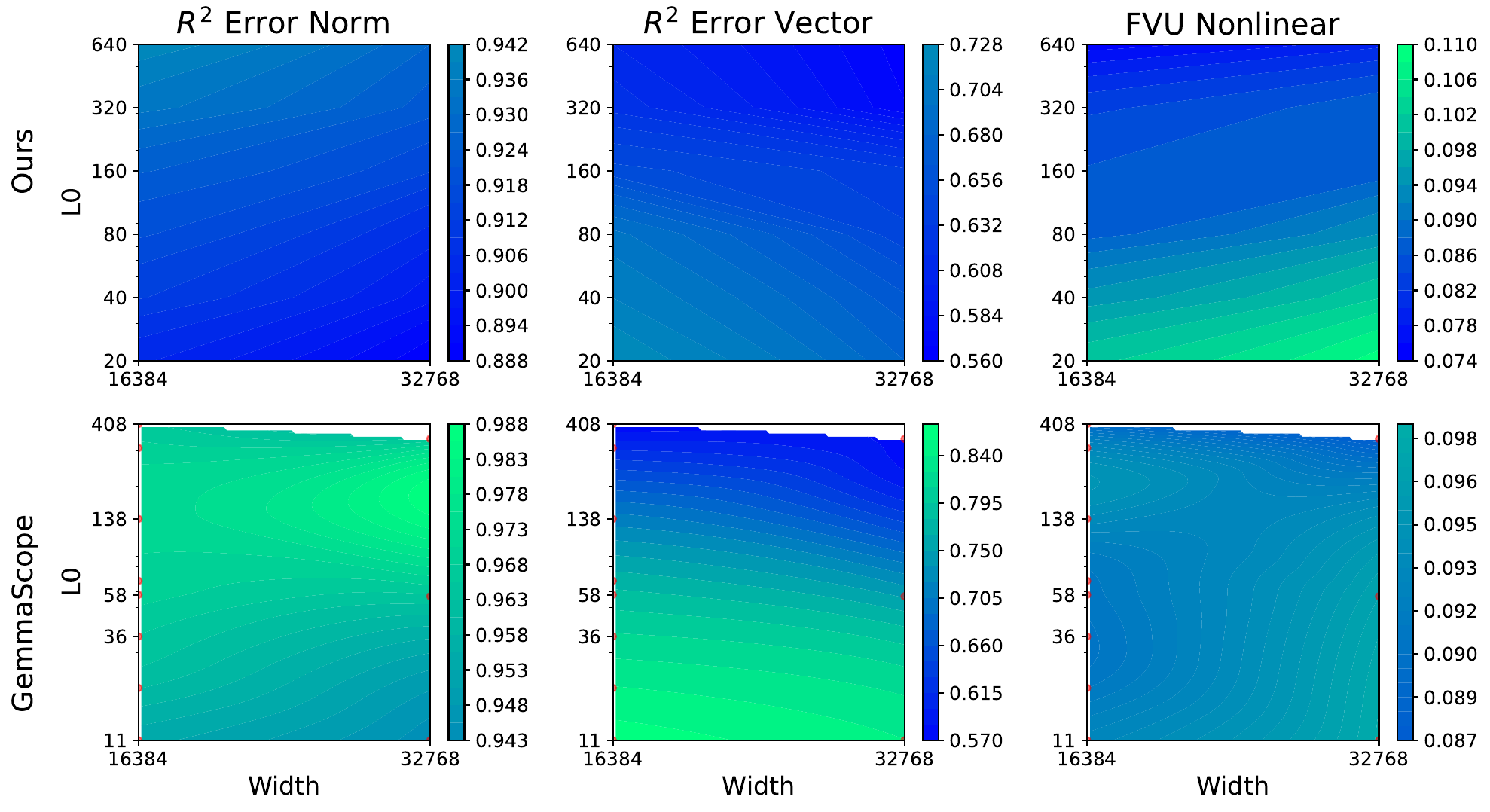}
    \caption{\textbf{Dark matter analysis} of reconstruction error for domain-specific SAEs (top row) vs.\ GemmaScope SAEs (bottom row), measured at layer~20 of Gemma-2-9b. \textbf{Left:}~\(R^2\) of a scalar probe predicting the error norm from \(\mathbf{x}\). \textbf{Centre:}~\(R^2\) of a linear map predicting the entire error vector. \textbf{Right:}~Fraction of variance unexplained (\(\mathrm{FVU}_{\mathrm{nonlinear}}\)) after combining the SAE reconstruction with the best linear approximation of its error.}
    \label{fig:dark_matter_comparison}
\end{figure}

Across different widths and sparsity levels for Gemma-2-9b, we consistently observe that our \textbf{domain-specific} SAEs (top row of Figure~\ref{fig:dark_matter_comparison}) achieve \textit{lower error norm predictability}, \textit{lower linear predictability} of the error itself (i.e.\ \(R^2\) error vector is not substantially inflated), and \textit{higher nonlinear fraction} (\(\mathrm{FVU}_{\mathrm{nonlinear}}\)) in the residual.

In other words, once our domain-specific SAEs capture the main \emph{linear} features relevant to the clinical text domain, the \emph{remaining} error is smaller in magnitude but more purely nonlinear. By contrast, the GemmaScope SAEs (bottom row, Figure~\ref{fig:dark_matter_comparison}) tend to leave behind more linearly predictable ``dark matter'' -- consistent with the interpretation that broad-domain SAEs underutilise their capacity for fine-grained domain features. Put differently, GemmaScope’s residual still contains substantial \emph{linear} structure that could, in principle, have been represented in the SAE’s dictionary, whereas our domain-specific SAEs allocate capacity to those linear patterns and thus push most leftover error into a harder-to-capture (and smaller) nonlinear remainder.

\section{Related Work}

\textit{Failure Modes of SAEs.} SAEs suffer from substantial reconstruction errors that degrade model performance. Inserting a 16M-latent SAE into GPT-4 resulted in a language modelling loss equivalent to a model trained on only 10\% of GPT-4’s compute \citep{gao2024scaling}, while using SAE reconstructions in GPT-2 small led to performance drops of 10\% on task-specific data and 40\% on general data \citep{makelov2024principledevaluationssparseautoencoders}. Expanding dictionary size and sparsity can mitigate errors but introduces computational costs and compromises interpretability by creating near one-to-one latent mappings. Sparse dictionary learning (SDL) methods with ``error nodes'' \citep{marks2024sparse} offer partial improvements, yet much of the reconstruction error remains linearly predictable, indicating unlearned structured features \citep{engels2024decomposing}. 

Despite their design, SAEs do not guarantee interpretable latents. Feature splitting, absorption, and compositional artifacts \citep{chanin2024absorption, till_2024_saes_true_features} suggest that excessive sparsity constraints can distort feature representations. Alternative optimisation objectives, such as minimising description length, may yield more meaningful decompositions \citep{ayonrinde2024interpretabilitycompressionreconsideringsae}. Furthermore, SAEs trained on pretraining data often fail to capture task-specific latents, missing key functional concepts needed for downstream applications. For instance, SAEs trained on pretraining corpora fail to encode features for refusal behaviour, while those trained on chat data do \citep{SAEsAreHighlyDatasetDependent}. These issues highlight the broader limitations of SAEs in mechanistic interpretability, as outlined in \citet{sharkey2025open}.

\noindent \textit{Domain-Specific SAEs.} SAEs have been applied across diverse domains beyond LLMs, uncovering interpretable features in genomics, proteomics, and neuroscience. In Evo~2, Batch‐TopK SAEs trained on layer-26 activations reveal biologically meaningful genomic features like exon–intron boundaries and transcription factor binding motifs~\citep{bussmann2024batchtopksparseautoencoders, Brixi2024}. Similarly, SAEs applied to ESM-2 embeddings extract latent dimensions corresponding to protein binding sites and structural motifs~\citep{simon2024interplm, garcia2025interpreting}. In AI safety, SAEs enable controlled knowledge removal by selectively downscaling activations of biology-related concepts~\citep{farrell2024applyingsparseautoencodersunlearn}. Genomic applications extend to sparse convolutional denoising autoencoders for linkage disequilibrium detection~\citep{chen2019sparse} and stacked SAE architectures for cancer classification from gene expression data~\citep{zenbout2020stacked}. SAEs also show promise in network anomaly detection~\citep{mazadu2022improved}, radiology diagnosis using vision transformers~\citep{abdulaal2024xrayworth15features}, and scientific literature analysis with neural embedding models~\citep{oneill2024disentanglingdenseembeddingssparse}. In neuroscience, SAEs have been used to model neuronal responses in the visual cortex~\citep{geadah2024sparse}, analyse neural recordings~\citep{almuqhim2021asd, theodosis2023learning}, and extract interpretable features from visual neurons~\citep{klindt2023identifyinginterpretablevisualfeatures}.

\section{Discussion}

In this work, we have demonstrated that domain --- specific SAEs—trained exclusively on medical text --- yield markedly improved performance compared to broad-domain, foundational SAEs. Our experiments show that these SAEs achieve 15–20\% higher fraction of variance explained, superior loss recovery, and improved cosine similarity between activations and reconstructions. Furthermore, our interpretability analyses reveal that the latent features extracted by domain-specific SAEs align more closely with clinically meaningful concepts, while dark matter analysis indicates that our models capture the majority of linear structure, leaving a smaller and more nonlinear residual. Finally, our steering evaluations confirm that interventions based on these latent features lead to more precise and effective control over domain-specific model behaviour.

A detailed examination of our results reinforces these findings. In unsupervised evaluations, the enhanced reconstruction fidelity of domain-specific SAEs is evidenced by increased variance explained and loss recovery, which are critical for faithful mechanistic interpretability. The interpretability pipeline further shows that the latent features not only carry higher automated F1 scores but also encapsulate nuanced clinical phenomena (e.g., taste sensations, diagnostic specificity, and mononucleosis) more robustly than features from broad-domain models. Our dark matter analysis reveals that while foundational SAEs leave behind a significant fraction of linearly predictable error \citep{engels2024decomposing}, our domain-specific models more effectively allocate capacity to capturing linear structure, relegating residual errors to a smaller, predominantly nonlinear component. In addition, steering experiments demonstrate that these refined latent features facilitate more targeted interventions, enhancing the model’s controllability—a promising sign for downstream applications in clinical settings.

Several failure modes in broad-domain SAEs further underscore the necessity of domain-specific training. Feature splitting and absorption, for example, are exacerbated in wider SAEs, where latents often fragment into overly specialised components or absorb token-aligned signals, reducing their interpretability \citep{chanin2024absorption, till_2024_saes_true_features}. This is likely driven by the sparsity penalty, which, as argued by \citet{anders_etal_2024_composedtoymodels_2d}, biases SAEs toward learning frequent feature compositions rather than atomic, semantically distinct latents. Additionally, attempts to mitigate reconstruction errors by increasing dictionary size—while effective in broad SAEs—lead to computationally expensive models that approach one-to-one mappings between latents and activations, undermining their usefulness for mechanistic interpretability \citep{gao2024scaling, makelov2024principledevaluationssparseautoencoders}. Furthermore, broad SAEs trained on pretraining distributions frequently fail to capture task-critical concepts, such as refusal mechanisms in chat models \citep{SAEsAreHighlyDatasetDependent}, highlighting the misalignment between general-purpose training and specialised interpretability needs. 

As discussed in \citet{sharkey2025open}, these issues are compounded by evaluation methodologies that rely on generic behavioural probes rather than assessing the meaningfulness of individual features in specific domains. For instance, SAEBench evaluates feature absorption by using features for ``word starts with \texttt{x}'', which is not useful for evaluating domain-specific feature absorption. 

Domain-specific SAEs force the model to allocate its latent capacity to high-fidelity, task-relevant features, mitigating feature fragmentation, improving reconstruction fidelity, and enhancing interpretability. These results suggest that shifting from a foundation-model scaling paradigm to domain-constrained SAE training is crucial for achieving accurate and actionable decompositions of model activations.

\subsection{Future Work}

Looking ahead, several promising avenues remain for extending this research. One primary direction is to scale our approach by training SAEs on a vastly larger corpus of medical text—including bioarXiv papers, extensive clinical textbooks, and upwards of 2 billion medicine-specific tokens—to further enhance feature granularity and robustness. Additionally, exploring methodological extensions such as alternative optimisation targets (e.g., minimising description length rather than solely enforcing sparsity \citep{ayonrinde2024interpretabilitycompressionreconsideringsae}) may yield even more interpretable latent decompositions. There is also significant potential in applying domain-specific SAE methods to other modalities or in the context of crosscoders, which could broaden the impact of our findings on mechanistic interpretability. Finally, further experiments are needed to assess the downstream benefits of these refined latent features --- such as in the construction of causal subcircuits using sparse feature circuits \citep{marks2024sparse} --- to validate their utility in real-world applications.

\bibliography{iclr2025_conference}

\appendix

\tableofcontents

\section{Derivation of the Expected Loss Derivative}
\label{app:expected-loss-derivative}

Here we derive the gradient of the expected loss with respect to the threshold parameter $\theta_i$. Starting from
\[
\mathcal{L}_\theta(\mathbf{x}) = \|\mathbf{x} - \hat{\mathbf{x}}(\mathbf{f}(\mathbf{x}))\|_2^2 + \lambda \sum_{i=1}^{M} H\big(\pi_i(\mathbf{x}) - \theta_i\big),
\]
one may use the Leibniz integral rule and properties of the Dirac delta function to show that
\begin{equation}
\frac{\partial \mathbb{E}_\mathbf{x}\big[\mathcal{L}_\theta(\mathbf{x})\big]}{\partial \theta_i} 
= \Big(\mathbb{E}_\mathbf{x}\big[I_i(\mathbf{x})\mid \pi_i(\mathbf{x}) = \theta_i\big] - \lambda\Big)p_i(\theta_i),
\label{eq:exact-derivative-app}
\end{equation}
where
\[
I_i(\mathbf{x}) := 2\theta_i\,\mathbf{d}_i\cdot\Big(\mathbf{x} - \hat{\mathbf{x}}(\mathbf{f}(\mathbf{x}))\Big),
\]
and $p_i(\theta_i)$ is the probability density of the pre-activation $\pi_i(\mathbf{x})$ at $\theta_i$.

\section{STE Pseudo-Derivatives and the KDE Connection}
\label{app:ste-kde-connection}

This appendix explains how the pseudo-derivatives defined in Equations \ref{eq:jr-ste} and \ref{eq:step-ste} naturally yield a kernel density estimator (KDE) for the gradient of the expected loss with respect to the threshold parameter \(\theta_i\).

Recall that when a function is discontinuous (as in the case of the Heaviside step function), standard backpropagation fails to provide meaningful gradient information. To address this, we replace the true derivative with a pseudo-derivative defined via a kernel function \(K\). In particular, the pseudo-derivative for the JumpReLU function is given by:
\[
\frac{\partial}{\partial \theta}\mathrm{JumpReLU}_{\theta}(z) \approx -\frac{\theta}{\varepsilon}\,K\!\Big(\frac{z-\theta}{\varepsilon}\Big),
\]
and similarly for the Heaviside function,
\[
\frac{\partial}{\partial \theta}H(z-\theta) \approx -\frac{1}{\varepsilon}\,K\!\Big(\frac{z-\theta}{\varepsilon}\Big),
\]
where \(\varepsilon > 0\) is a small bandwidth parameter.

The kernel \(K\) serves as a smooth approximation to the Dirac delta function. Its role is to ``smear'' the discontinuity over a small interval, allowing gradients to propagate through points where the original function is not differentiable.

In kernel density estimation, given samples \(x_1, \dots, x_N\), the density at a point \(x\) is estimated as:
\begin{equation}
\hat{p}_X(x) = \frac{1}{N\varepsilon} \sum_{\alpha=1}^{N} K\!\Big(\frac{x-x_\alpha}{\varepsilon}\Big).
\label{eq:kde-estimator-app}
\end{equation}

In our setting, the gradient of the expected loss with respect to \(\theta_i\) involves contributions from the reconstruction error and the sparsity penalty. For each sample \(\mathbf{x}_\alpha\), these contributions are modulated by the kernel \(K\!\Big(\frac{\pi_i(\mathbf{x}_\alpha)-\theta_i}{\varepsilon}\Big)\), which localises the gradient estimation around the threshold \(\theta_i\). Concretely, the batch-wise gradient approximation is given by:
\begin{equation}
\frac{1}{N\varepsilon}\sum_{\alpha=1}^{N} \Big[I_i(\mathbf{x}_\alpha) - \lambda\Big]\, K\!\Big(\frac{\pi_i(\mathbf{x}_\alpha)-\theta_i}{\varepsilon}\Big),
\label{eq:kde-estimate-app}
\end{equation}
where
\[
I_i(\mathbf{x}_\alpha) := 2\theta_i\,\mathbf{d}_i\cdot\Big(\mathbf{x}_\alpha - \hat{\mathbf{x}}(\mathbf{f}(\mathbf{x}_\alpha))\Big)
\]
captures the contribution from the reconstruction loss, and \(\lambda\) scales the contribution from the sparsity penalty.

This expression is analogous to the KDE estimator in Equation~\ref{eq:kde-estimator-app}, but with each sample weighted by the term \(\Big[I_i(\mathbf{x}_\alpha) - \lambda\Big]\). In other words, the STE-based gradient calculation performs a local averaging (or smoothing) over the mini-batch, effectively estimating the gradient by aggregating contributions from samples whose pre-activations \(\pi_i(\mathbf{x}_\alpha)\) lie near the threshold \(\theta_i\).

Alternative kernel choices (such as triangular, Gaussian, or Epanechnikov kernels) can be used in place of the one presented here, and they all yield a similar smoothing effect in the gradient estimation. This connection to KDE provides an intuitive interpretation of the STE: it estimates the gradient of the expected loss by effectively performing a density-weighted average of the sample-wise contributions near the point of interest.

\subsection{Probability-Based Intuition}
\label{app:probability-intuition}

Another perspective is to view the activations as random variables. The expected L0 penalty is given by:
\[
\mathbb{E}_\mathbf{x}\|\mathbf{f}(\mathbf{x})\|_0 = \sum_{i=1}^{M} \mathbb{P}\big(\pi_i(\mathbf{x}) > \theta_i\big),
\]
which is differentiable with respect to $\theta_i$. In fact, 
\[
\frac{d}{d\theta_i}\mathbb{P}\big(\pi_i(\mathbf{x}) > \theta_i\big) = -p_i(\theta_i).
\]
Thus, the STE approximates this derivative by replacing the Dirac delta function with a smoothed kernel \(K\). This probabilistic interpretation underlies the KDE connection discussed in Appendix~\ref{app:ste-kde-connection}.

\section{Derivation of the Relative Reconstruction Bias Metric}
\label{app:reconstruction-bias}

We aim to determine the optimal multiplicative factor 
\[
\gamma := \underset{\gamma'}{\arg\min}\; \mathbb{E}_{\mathbf{x} \sim \mathcal{D}}\Big[\Big\|\frac{\hat{\mathbf{x}}_{\mathrm{SAE}}(\mathbf{x})}{\gamma'} - \mathbf{x}\Big\|_2^2\Big],
\]
where \(\hat{\mathbf{x}}_{\mathrm{SAE}}(\mathbf{x})\) is the SAE’s reconstruction of \(\mathbf{x}\). Define the error function:
\[
E(\gamma') = \mathbb{E}_{\mathbf{x} \sim \mathcal{D}}\Big[\Big\|\frac{\hat{\mathbf{x}}_{\mathrm{SAE}}(\mathbf{x})}{\gamma'} - \mathbf{x}\Big\|_2^2\Big].
\]
For each \(\mathbf{x}\), expanding the squared L2 norm gives:
\[
\left\|\frac{\hat{\mathbf{x}}_{\mathrm{SAE}}(\mathbf{x})}{\gamma'} - \mathbf{x}\right\|_2^2 
=\frac{1}{\gamma'^2}\|\hat{\mathbf{x}}_{\mathrm{SAE}}(\mathbf{x})\|_2^2 - \frac{2}{\gamma'}\hat{\mathbf{x}}_{\mathrm{SAE}}(\mathbf{x})\cdot\mathbf{x} + \|\mathbf{x}\|_2^2.
\]
Define
\[
A = \mathbb{E}\Big[\|\hat{\mathbf{x}}_{\mathrm{SAE}}(\mathbf{x})\|_2^2\Big], \quad B = \mathbb{E}\Big[\hat{\mathbf{x}}_{\mathrm{SAE}}(\mathbf{x})\cdot\mathbf{x}\Big], \quad C = \mathbb{E}\Big[\|\mathbf{x}\|_2^2\Big].
\]
Thus,
\[
E(\gamma') = \frac{A}{\gamma'^2} - \frac{2B}{\gamma'} + C.
\]
Minimising \(E(\gamma')\) with respect to \(\gamma'\), we differentiate:
\[
\frac{dE}{d\gamma'} = -\frac{2A}{\gamma'^3} + \frac{2B}{\gamma'^2} = 0.
\]
Multiplying by \(\gamma'^3/2\) (which is positive) yields:
\[
-A + B\gamma' = 0 \quad\Longrightarrow\quad \gamma' = \frac{A}{B}.
\]
Hence, the optimal scaling factor is:
\[
\gamma = \frac{A}{B} = \frac{\mathbb{E}\big[\|\hat{\mathbf{x}}_{\mathrm{SAE}}(\mathbf{x})\|_2^2\big]}{\mathbb{E}\big[\hat{\mathbf{x}}_{\mathrm{SAE}}(\mathbf{x})\cdot\mathbf{x}\big]}.
\]
We now express \(B\) in terms of the mean squared reconstruction error. Noting that
\[
\|\hat{\mathbf{x}}_{\mathrm{SAE}}(\mathbf{x}) - \mathbf{x}\|_2^2 = \|\hat{\mathbf{x}}_{\mathrm{SAE}}(\mathbf{x})\|_2^2 + \|\mathbf{x}\|_2^2 - 2\hat{\mathbf{x}}_{\mathrm{SAE}}(\mathbf{x})\cdot\mathbf{x},
\]
taking expectations gives:
\[
D = \mathbb{E}\Big[\|\hat{\mathbf{x}}_{\mathrm{SAE}}(\mathbf{x}) - \mathbf{x}\|_2^2\Big] = A + C - 2B.
\]
Solving for \(B\):
\[
B = \frac{A + C - D}{2}.
\]
Substituting back, we obtain:
\[
\gamma = \frac{A}{\frac{A + C - D}{2}} = \frac{2A}{A + C - D}.
\]
Thus, the relative reconstruction bias is given by
\[
\gamma = \frac{2\,\mathbb{E}\big[\|\hat{\mathbf{x}}_{\mathrm{SAE}}(\mathbf{x})\|_2^2\big]}{\mathbb{E}\big[\|\hat{\mathbf{x}}_{\mathrm{SAE}}(\mathbf{x})\|_2^2\big] + \mathbb{E}\big[\|\mathbf{x}\|_2^2\big] - \mathbb{E}\big[\|\hat{\mathbf{x}}_{\mathrm{SAE}}(\mathbf{x}) - \mathbf{x}\|_2^2\big]}.
\]
\textbf{Additional Remarks:}  
The derivation uses the identity
\[
2\,\mathbf{a}\cdot\mathbf{b} = \|\mathbf{a}\|_2^2 + \|\mathbf{b}\|_2^2 - \|\mathbf{a}-\mathbf{b}\|_2^2.
\]
An unbiased reconstruction (perfect SAE) yields \(\gamma = 1\); however, in practice, L1 regularization may induce shrinkage (\(\gamma < 1\)). This metric thus quantifies the extent to which the SAE's decoder underestimates the input norm.

\section{Automated Intepretability Details}
\label{app:autointerp}

In our evaluation, we employ the unified \texttt{charlieoneill/medical-qa-combined} dataset, processing a total of 49 million tokens. Each activation context is constrained to 256 tokens, and the activations are processed in batches of 8 examples. For each latent feature, our sampling procedure ensures that only those with at least 250 examples are considered, while we cap the number of examples per latent at $5\,000$. During training, we sample 40 examples per latent using a quantile-based method, and for evaluation, 50 test examples per latent are selected. In addition, non-activating examples are randomly sampled from contexts that do not trigger the latent, which provides a balanced set of inputs for our detection pipeline.

To limit the computational load, our evaluation is restricted to a maximum of $1\,000$ latent features. The LLM explainer is configured to use OpenAI's \texttt{gpt-4o-mini model}, with the maximum context length set to 4208 tokens. This setting ensures that the explainer has sufficient context to generate high-quality explanations without exceeding model limitations.

The entire evaluation pipeline is implemented asynchronously to manage the large-scale data processing efficiently. This asynchronous design spans activation caching, latent example construction, and both detection and fuzzing scoring. 

\section{Additional SAE Evaluations}
\label{app:additional_sae_evaluations}

\begin{figure}
    \centering
    \includegraphics[width=0.98\linewidth]{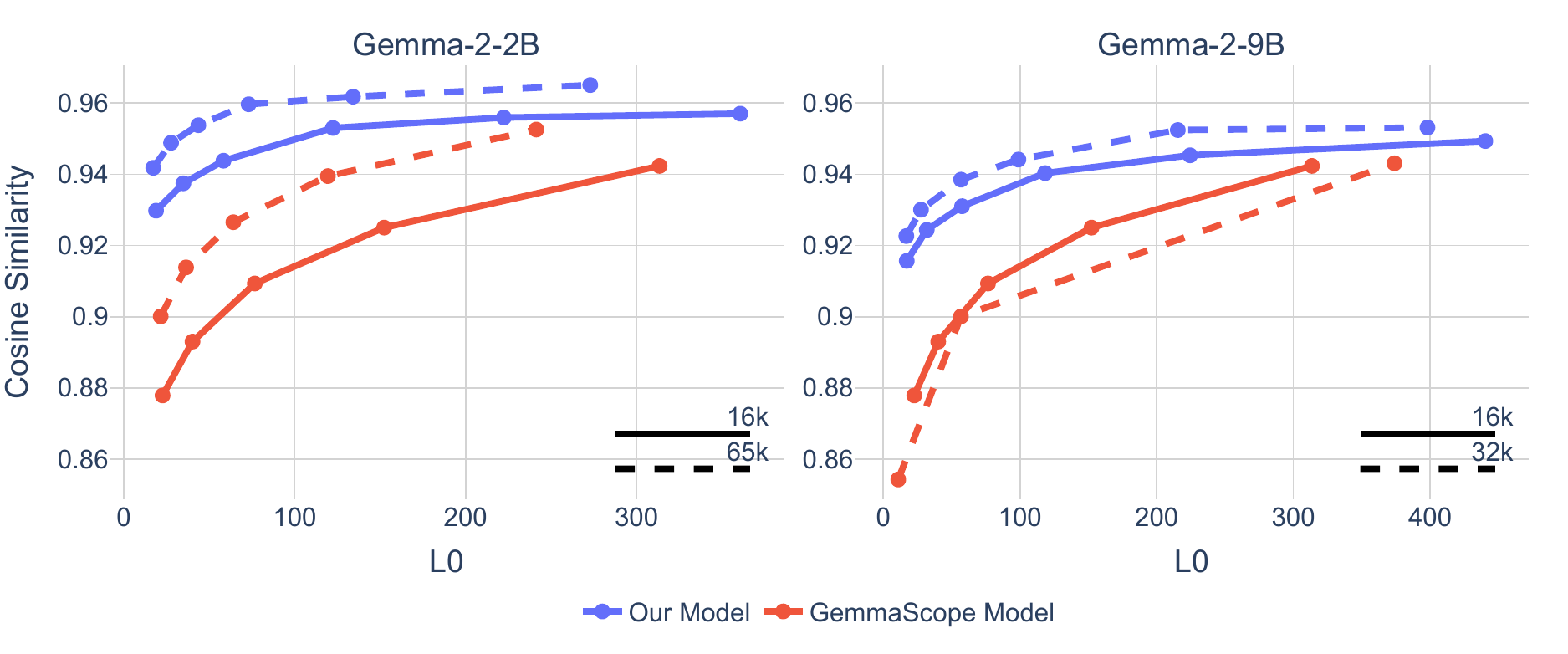}
    \caption{Pareto curves of cosine similarity, between the model activations and SAE reconstructions, across different models (\texttt{gemma-2-2b} and \texttt{gemma-2-9b}), SAE widths, and sparsities, comparing our SAEs to the GemmaScope SAEs.}
    \label{fig:cossim}
\end{figure}

\begin{figure}
    \centering
    \includegraphics[width=0.98\linewidth]{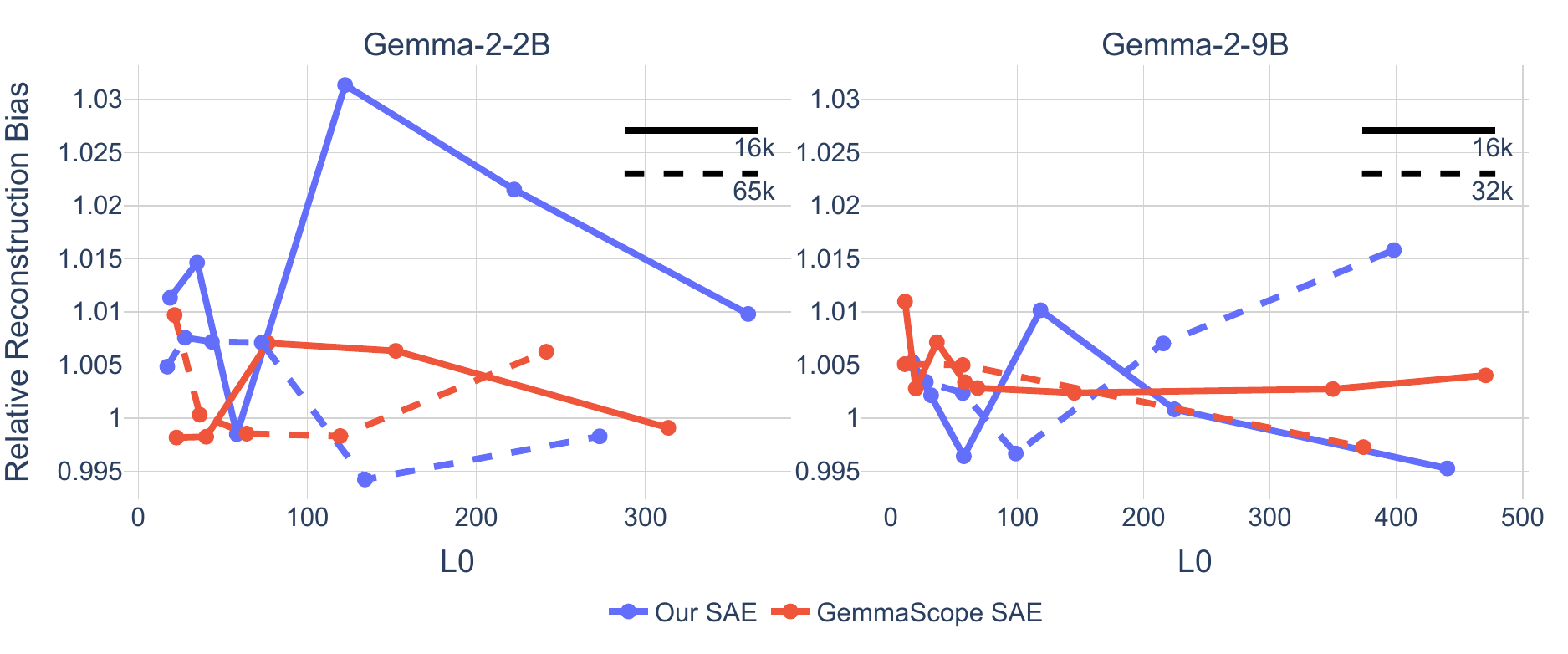}
    \caption{Pareto curves of relative reconstruction bias across different models (\texttt{gemma-2-2b} and \texttt{gemma-2-9b}), SAE widths, and sparsities, comparing our SAEs to the GemmaScope SAEs.}
    \label{fig:relative_reconstruction_bias}
\end{figure}

\section{Feature Similarity Analysis}
\label{app:feature-sim}

\begin{figure}
    \centering
    \includegraphics[width=0.5\linewidth]{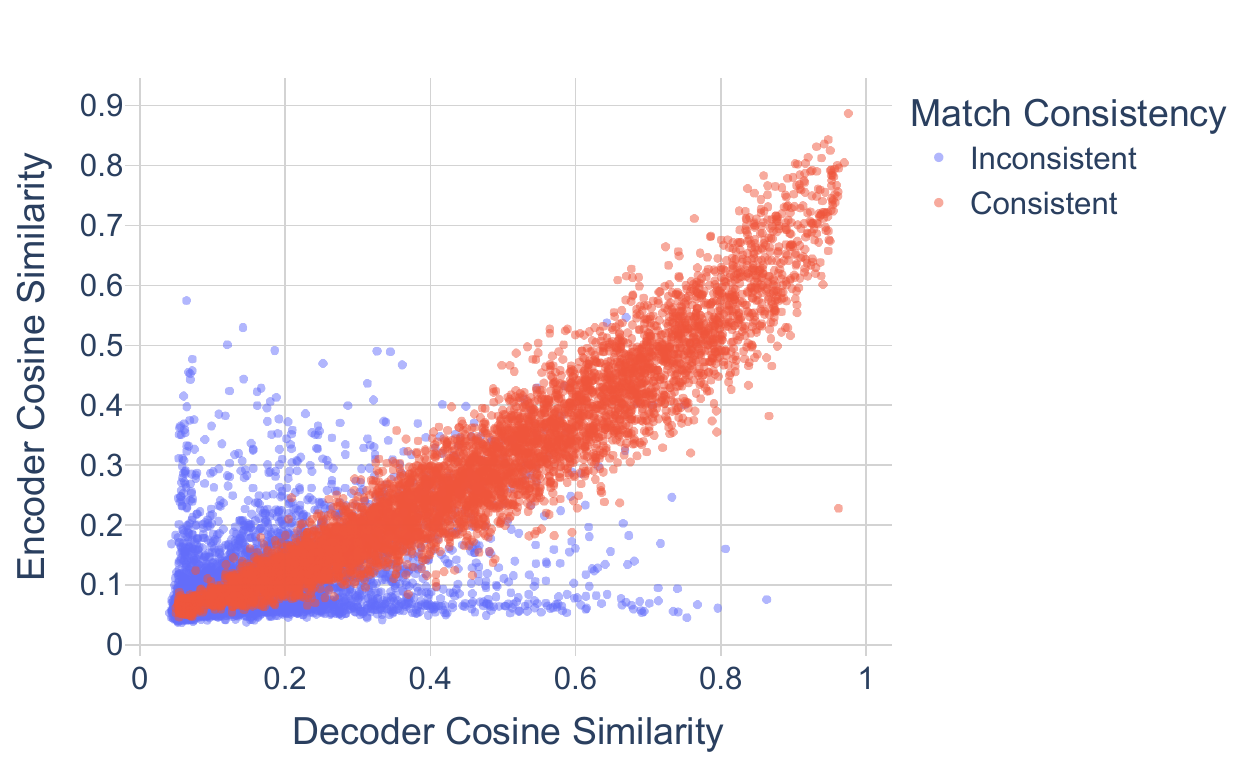}
    \caption{Matched feature similarities for Gemma and our SAE decoders and encoders. Each point represents a Gemma feature vector, with its matched cosine similarity in decoder space on the x-axis and encoder space on the y-axis. Points are coloured by whether the matched index in our SAE is consistent for both decoder and encoder.}
    \label{fig:decoder_encoder_similarity_hungarian}
\end{figure}

We applied the Hungarian algorithm to match each feature vector from Gemma’s decoder and encoder to the nearest feature vector in our SAE based on cosine similarity \citep{paulo2025sparse, oneill2025computeoptimalinferenceprovable}. Each point in Figure~\ref{fig:decoder_encoder_similarity_hungarian} corresponds to one GemmaScope feature, with the x-axis showing the cosine similarity in the decoder space and the y-axis showing the cosine similarity in the encoder space. Points are coloured according to whether both matches (decoder and encoder) mapped to the same index in our SAE. Most points with high decoder similarity also have high encoder similarity, suggesting that these features align consistently in both representations. A smaller group of inconsistent matches indicates features that do not map to the same index across the two spaces.

\begin{figure}
    \centering
    \includegraphics[width=0.7\linewidth]{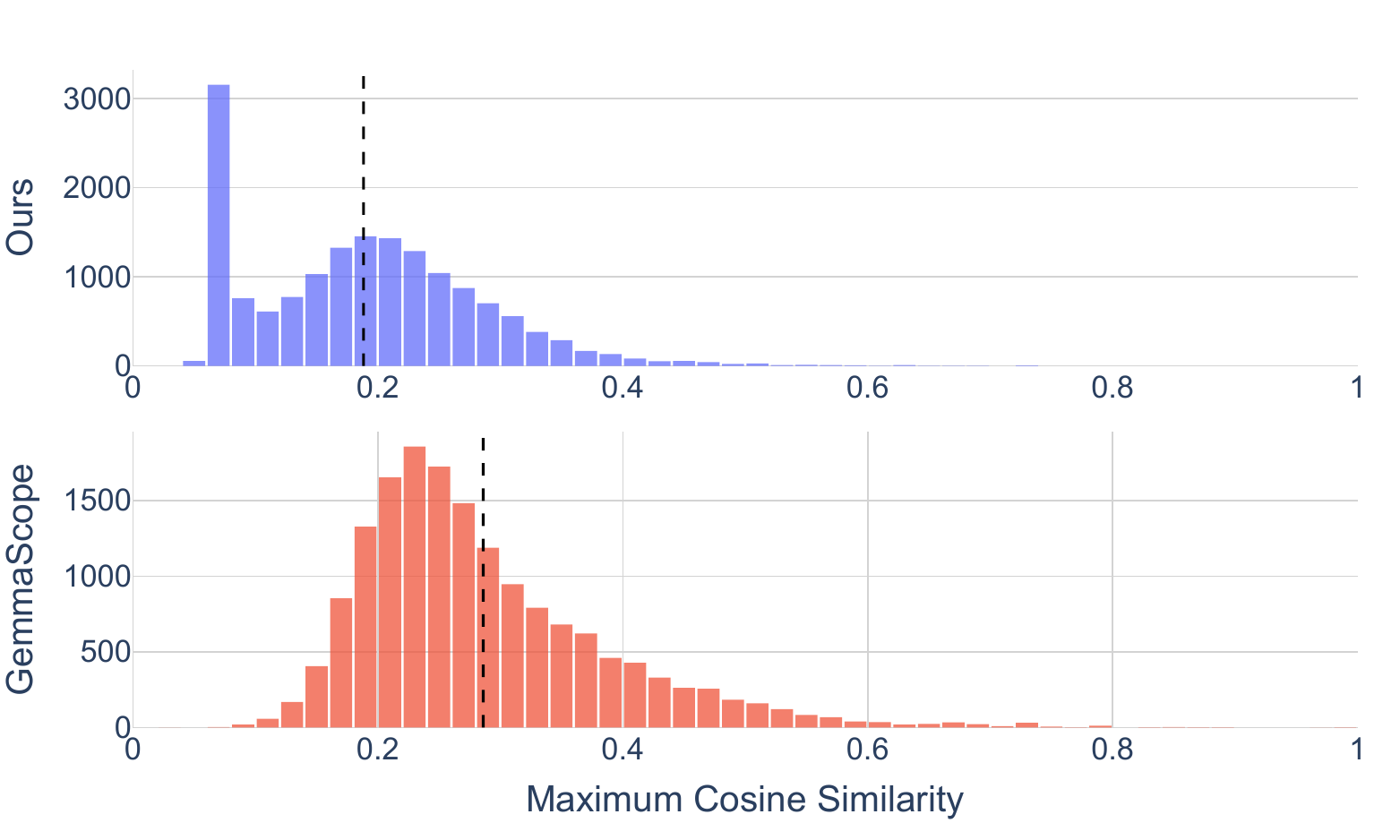}
    \caption{Distribution of maximum cosine similarities for the learned SAE features}
    \label{fig:intersim_histogram}
\end{figure}

\end{document}

%% file: math_commands.tex
\usepackage{amsmath,amsfonts,bm}

\def\eqref#1{equation~\ref{#1}}

\def\1{\bm{1}}

\DeclareMathAlphabet{\mathsfit}{\encodingdefault}{\sfdefault}{m}{sl}
\SetMathAlphabet{\mathsfit}{bold}{\encodingdefault}{\sfdefault}{bx}{n}

